\documentclass[letterpaper, 10 pt, conference]{ieeeconf}  

\IEEEoverridecommandlockouts    

\usepackage[printonlyused,nohyperlinks]{acronym}
\usepackage{subcaption}

\usepackage{algorithm}
\usepackage{algorithmicx}
\usepackage[noend]{algpseudocode}
\usepackage{etoolbox}
\usepackage{amssymb}
\usepackage{amsmath}
\usepackage{svg}
\usepackage{xspace}
\usepackage{booktabs}
\usepackage{float}

\usepackage{graphicx}
\usepackage{multirow}
\usepackage{verbatim}

\usepackage{array}
\usepackage{adjustbox}
\usepackage{wrapfig}

\usepackage{tikz}
\usetikzlibrary{shapes, arrows, positioning, fit, backgrounds}
\usetikzlibrary{calc}
\usetikzlibrary{decorations.pathmorphing}
\usepackage{caption}
\usepackage{siunitx}
\usepackage[pagebackref=true,breaklinks=true,colorlinks,bookmarks=false,citecolor=blue]{hyperref}     
\usepackage{url}  

\usepackage[T1]{fontenc}
\usepackage[utf8]{inputenc}

\tikzstyle{block} = [rectangle, draw, fill=blue!20, 
    text centered, rounded corners, minimum height=1em, node distance=1.5cm]
\tikzstyle{line} = [draw, -latex']
\tikzstyle{cloud} = [draw, rectangle,fill=red!20, node distance=1.5cm and 1cm,
    minimum height=1em]

\usepackage{array,multirow}
\usepackage{cprotect}
\usepackage{csquotes}

\algrenewcommand\algorithmicindent{0.5em}%

\usetikzlibrary{tikzmark}
\usetikzlibrary{calc}

\errorcontextlines\maxdimen

\newcommand{\ALGtikzmarkcolor}{black}
\newcommand{\ALGtikzmarkextraindent}{2pt}
\newcommand{\ALGtikzmarkverticaloffsetstart}{-.5ex}
\newcommand{\ALGtikzmarkverticaloffsetend}{-.5ex}
\makeatletter
\newcounter{ALG@tikzmark@tempcnta}

\newcommand\ALG@tikzmark@start{%
    \global\let\ALG@tikzmark@last\ALG@tikzmark@starttext%
    \expandafter\edef\csname ALG@tikzmark@\theALG@nested\endcsname{\theALG@tikzmark@tempcnta}%
    \tikzmark{ALG@tikzmark@start@\csname ALG@tikzmark@\theALG@nested\endcsname}%
    \addtocounter{ALG@tikzmark@tempcnta}{1}%
}

\def\ALG@tikzmark@starttext{start}
\newcommand\ALG@tikzmark@end{%
    \ifx\ALG@tikzmark@last\ALG@tikzmark@starttext
    \else
        \tikzmark{ALG@tikzmark@end@\csname ALG@tikzmark@\theALG@nested\endcsname}%
        \tikz[overlay,remember picture] \draw[\ALGtikzmarkcolor] let \p{S}=($(pic cs:ALG@tikzmark@start@\csname ALG@tikzmark@\theALG@nested\endcsname)+(\ALGtikzmarkextraindent,\ALGtikzmarkverticaloffsetstart)$), \p{E}=($(pic cs:ALG@tikzmark@end@\csname ALG@tikzmark@\theALG@nested\endcsname)+(\ALGtikzmarkextraindent,\ALGtikzmarkverticaloffsetend)$) in (\x{S},\y{S})--(\x{S},\y{E});%
    \fi
    \gdef\ALG@tikzmark@last{end}%
}

\newcommand{\videourl}{\url{https://taochenshh.github.io/projects/tactofind}}

\apptocmd{\ALG@beginblock}{\ALG@tikzmark@start}{}{\errmessage{failed to patch}}
\pretocmd{\ALG@endblock}{\ALG@tikzmark@end}{}{\errmessage{failed to patch}}
\makeatother
\makeatletter
\def\thanks#1{\protected@xdef\@thanks{\@thanks
        \protect\footnotetext{#1}}}
\makeatother

\title{\LARGE \bf
TactoFind: A Tactile Only System for Object Retrieval 
}
\author{Sameer Pai$^{*,1,2}$, Tao Chen$^{*,1,2}$, Megha Tippur$^{*,2}$, Edward Adelson$^2$, Abhishek Gupta$^{\dagger,1,2, 3}$, Pulkit Agrawal$^{\dagger,1,2}$\\
$^1$Improbable AI Lab $\quad$ $^2$Massachusetts Institute of Technology $^3$University of Washington\\
$^*$Authors contributed equally. $^\dagger$Equal Advising. \\}

\begin{document}
\maketitle

\begin{abstract}
We study the problem of object retrieval in scenarios where visual sensing is absent, object shapes are unknown beforehand and objects can move freely, like grabbing objects out of a drawer. Successful solutions require localizing free objects, identifying specific object instances, and then grasping the identified objects, only using touch feedback. Unlike vision, where cameras can observe the entire scene, touch sensors are local and only observe parts of the scene that are in contact with the manipulator. Moreover, information gathering via touch sensors necessitates applying forces on the touched surface which may disturb the scene itself. Reasoning with touch, therefore, requires \textit{careful} exploration and integration of information over time -- a challenge we tackle.  
We present a system capable of using sparse tactile feedback from fingertip touch sensors on a dexterous hand to localize, identify and grasp novel objects without any visual feedback.  
Videos are available at \videourl.
\end{abstract}

\section{Introduction}
\label{sec:introduction}
\begin{figure}[!t]
    \centering
    \includegraphics[width=0.95\linewidth]{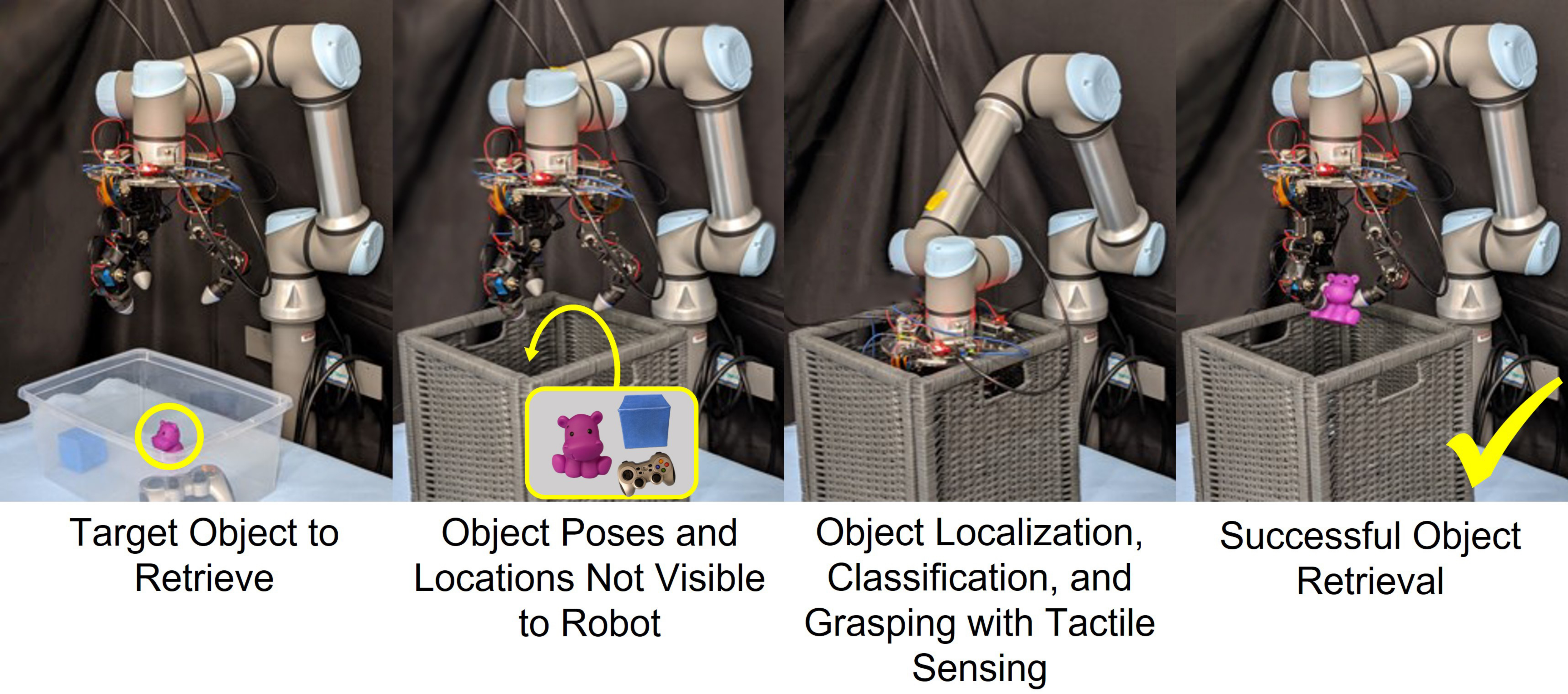}
    \caption{\footnotesize{System setup showing object retrieval from an occluded bin using only tactile sensing in the real world. The agent has to localize and identify the object from fingertip tactile sensors alone}}
    \label{fig:teaser}
    \vspace{-0.5cm}
\end{figure}

Consider the setup in Fig~\ref{fig:teaser} wherein multiple objects are placed in a box with unknown positions and orientations. A multi-fingered robot is tasked to fetch a particular object using only tactile observations from sensors placed on fingertips, without access to visual observations. Such problems are commonly encountered in daily life --- retrieving a desired object from the backpack, inside of a drawer, or from a tall cabinet where the topmost shelves are not visually observable. In these situations, the sense of touch can compensate for the lack of visual observations. 
To the best of our knowledge, the capability of a robotic system successfully retrieving a desired object based only on tactile observations has not been demonstrated in scenes with multiple freely moving and unknown objects. 

To understand what makes vision-free object retrieval challenging, first consider a scenario where the robot can visually observe the scene. In such a case, it is (typically) possible to identify the location and identity of all objects from just a single image. Furthermore, the same image communicates enough information about object geometry to plan a grasp. In contrast, tactile observations made by the fingertips are local and sparse. Fingertips only make infrequent contact with a local area of the object. Therefore the robot first needs to explore to localize objects. Then it needs to plan a sequence of touches on each object to gather enough information to identify and grasp it. Unlike visual data, where a single observation is usually sufficient, tactile-based object retrieval requires careful planning of a long sequence of actions for information gathering and mechanisms for temporal integration of this information. 

Usually, one would expect that as the robot interacts more with the environment, it gathers more ``information". However, this is not always true: generating touch readings requires applying force on an object that might move it. Such motion creates difficulty both in localization and identification. During localization, a moving object might contact another object previously encountered by the robot and potentially invalidate existing state estimates such as the object's location and pose. Similarly, during classification, if the object moves while being touched, the shape estimation is noisier which can make object identification difficult.

To summarize, the two key problems that make object retrieval in the absence of vision challenging are: \textbf{Firstly}, there is significant partial observability resulting from local and sparse touch observations. \textbf{Secondly}, obtaining touch readings requires force application on objects which may, in turn, lead to a change in object pose. While prior work has studied object classification and localization using tactile observations ~\cite{bottcher2021object, bauza2019tactile}, this has primarily been limited to single object scenes with a few classes of stationary objects. 

We present a system that overcomes these challenges and is capable of exploring and retrieving novel objects only from local fingertip tactile observations in the presence of several movable objects, albeit in a simplified setup where individual objects are separated. Our solution employs a localization scheme that minimizes object motion while clustering to find object positions. Once objects are localized, the robot performs directed exploration to infer object shape by making careful touches around the determined object location. The shape information gathered from the sequence of contacts is represented as a point cloud. We leverage self-supervised contrastive learning~\cite{chopra05contrastive, oord2018cpc} to embed tactile point clouds into a feature space used for object identification. To retrieve localized and identified objects, a simple grasping system is deployed. We present results on a 3-fingered robotic hand-arm system in both simulation and the real world. Our system achieves $76\%$ success in simulation and $60\%$ in the real world at retrieving novel objects without visual feedback.

\section{Related Work}
\label{sec:rw}
In recent years, much progress has been made in the development and usage of tactile sensors to improve a robot’s ability to complete dexterous manipulation tasks \cite{kappassov2015tactile}. In particular, sensors utilizing a variety of transduction methods, such as resistance \cite{sundaram2019learning} , capacitance \cite{boutry2018hierarchically}, piezo-electrical \cite{cutkosky2008force}, magnetic \cite{bhirangi2021reskin}, and optical \cite{ward2018tactip, kuppuswamy2020soft, lambeta2020digit}, have been shown to help robots more intelligently interact with the physical world \cite{lambeta2020digit, she2021cable, wang2020swingbot, dong2021tactile}. In our work, we leverage tactile sensing to perform object retrieval completely in the absence of visual input.

\subsection{Object Classification with Tactile Sensors}
 
While there has been much work in object classification using a combination of image and tactile information \cite{chaudhury2022using, yang2015object, corradi2017object}, our problem requires us to rely only on tactile data for classification. \cite{yuan2018active}, \cite{li2013sensing}, and \cite{kuppuswamy2020soft} classified different fabrics, textures, and basic geometries (respectively) using only camera-based tactile sensors. Others, such as \cite{bottcher2021object, pezzementi2011mapping, pohtongkam2021tactile}, used force-based tactile sensors to accomplish the classification task but often assumed the object was static. \cite{lin2019multimodal} aims to do tactile-based classification by leveraging multi-modal information across vision and touch. A class of methods also aim to make full shape inference from tactile feedback ~\cite{jamali2016activetactile, uriel2013contour,yi2016activeexploration, driess2017activetactile, xu2022tandem} but typically assume the object is static and there is only a single object in the scene. In contrast, in this work, we also operate in the tactile-only regime, but we interact with multiple non-static objects when collecting the surface touches used in classification and just look to make object identification, not ideal shape reconstruction. Moreover, we are able to operate in a regime where we do not require any explicit object models or knowledge of object dynamics at test time, making the method easy to apply broadly to a variety of novel objects. 

\subsection{Object Localization with Tactile Sensing}
Localization and manipulation of objects using only tactile sensing have been explored in prior work. While some works \cite{Karayiannidis2014OnlineCP, likar2014external} focused more on the dynamics and kinematics of the robot to perform localization,  Li et al. \cite{li2014localization} used a GelSight sensor to localize the sensed portion of single small objects like USB cables relative to the gripper. Bauza et al. \cite{bauza2019tactile} adopt a similar approach, combining heightmap information obtained from a tactile sensor with the robot’s kinematics and the iterated closest point method to localize and even identify small objects. Other works ~\cite{luo2017visualmap} approach the localization problem by also building a visual map, but do so assuming objects are fixed. ~\cite{koval2015manifold} perform contact-based localization to perform accurate manipulation when the object is in hand. In contrast to these works, we are less focused on fine-grained, precise, and in-hand localization than localization strategies that can deal with multiple, much larger movable objects while also performing identification and grasping. We compromise significantly on accuracy to obtain a general strategy that can approximately localize objects without requiring known object shapes and dynamics by aiming to keep the scene static. 

\cite{kissoum2021simultaneous} goes beyond static objects and uses a SLAM approach to do planar object reconstruction and localization of non-static objects in the plane in single object scenes, leveraging a Gaussian process implicit surface method and particle filtering. Methods like \cite{murali18blind, petrovskaya11localization} also attempted to use only tactile sensing to retrieve a non-static object in a single object scene using particle filtering methods. Additionally, ~\cite{kaboli2017pretouch} uses a pre-touch sensor to localize and then senses object properties like stiffness and sliding for object discrimination. In contrast, in our work, we interact with multiple movable objects with a dexterous 3-fingered manipulator and perform the whole pipeline of localization, identification \emph{and} grasping for object retrieval only using fingertip tactile sensors.

\begin{figure*}[!tb]
    \centering
    \includegraphics[width=1.\linewidth]{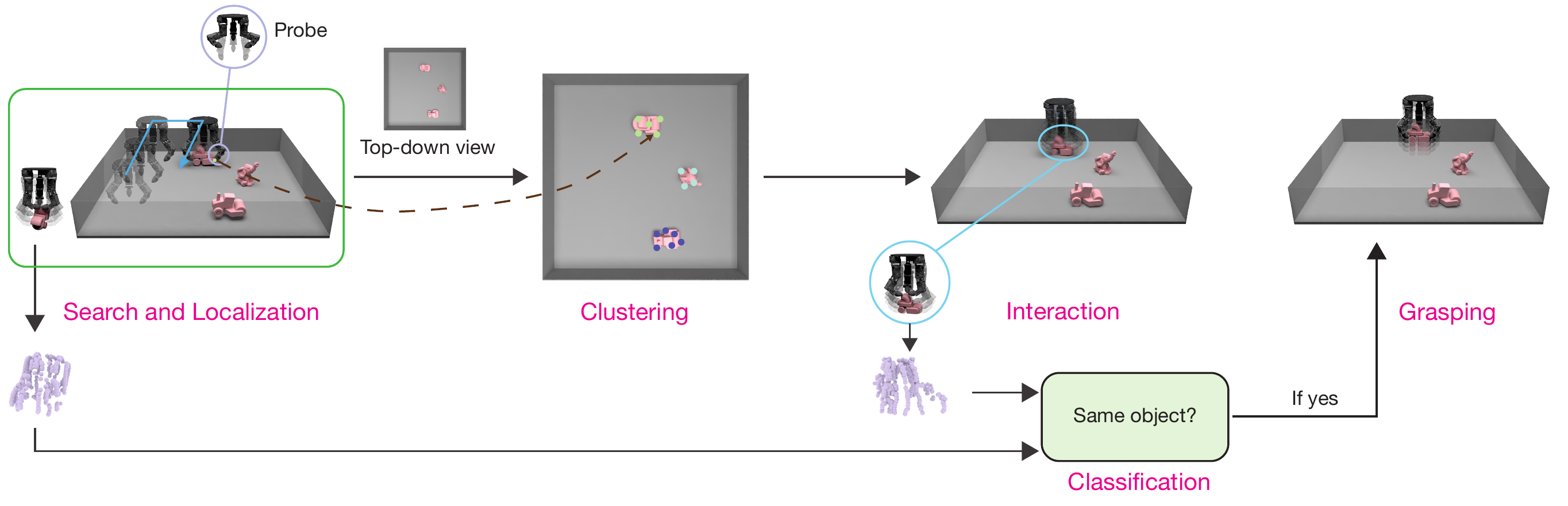}
    \caption{\footnotesize{Depiction of the full pipeline for tactile-only object retrieval. The agent first localizes several objects in the scene by vertically probing in a discrete grid around the environment, while applying minimal force. The object is then interacted with by radially tapping it gently for identification. The collected data is then used to identify if the object matches the object it is tasked to retrieve and then grasping is performed to actually retrieve the object.}}
    \label{fig:pipeline}
    \vspace{-0.5cm}
\end{figure*}

\section{System Description}
\label{sec:system}
Fig~\ref{fig:teaser} illustrates our system consisting of a three-fingered robotic hand mounted on a UR5e robotic arm (controlled by \cite{airobot2019}), equipped with fingertip GelSight 360 \cite{tippur2022design} for touch sensing. \textbf{Robotic Hand}: as shown in Fig~\ref{fig:teaser}, we use the three-fingered D'Claw robotic hand system~\cite{ahn2020robel}. Each finger has three Dynamixel servo motors connected in series for a total of nine degrees of freedom. The fingers are position controlled. \textbf{Tactile Sensor}:
we use omnidirectional GelSight sensors \cite{tippur2022design} on the tip of each finger of the D'Claw hand. These sensors contain fisheye camera lens surrounded by a soft elastomer gel that deforms when the sensor makes contact. The deformations are recorded as a RGB image by the camera. This GelSight sensor has a significantly larger area of contact (approximately $15$ cm\textsuperscript{$2$}) with objects than most sensors since it wraps completely around the finger. This is important to actually help identify objects with fewer interactions. To obtain binary contact readings from the sensor's raw data (RGB images), we train a force recognition model (as described in~\cite{sun2022soft}) in the real world to map images of contact from the GelSight sensor to binary labels of whether contact occurred or not.  

\section{Problem Setup}
\label{sec:prelims}
We first task the agent by allowing it to interact with a given \textit{target object} at a known location and pose, as shown in Fig~\ref{fig:pipeline}. The target object is then placed in a planar bin area along with other \textit{distractor objects}, each in a random unknown pose. The agent's goal is to leverage only tactile sensing to retrieve the target object. All objects are rigid and movable, and the agent has no prior knowledge of distractor object shapes. Furthermore, we assume no objects are stacked on each other for tractability. The robot observes fingertip positions $\{p_i\in\mathbb{R}^3, i=1,2,3\}$ estimated through joint encoders, as well as a reading of binary contact $\{c_i\in\{0, 1\}, i=1,2,3\}$ for each finger. 

\section{TactoFind: Identifying and Retrieving Objects in the absence of Visual Feedback}
\label{sec:method}
The problem of blind identification and retrieval of novel and movable objects is an instance of \textit{partially observable Markov decision process}, which is known to be intractable in general. To make this problem more approachable, instead of solving end-to-end, we can take a step-wise approach illustrated in Fig~\ref{fig:pipeline} and make assumptions detailed below. The robot first performs object localization, then instance identification, followed by grasping. In the \textit{localization phase}, the location of all objects is estimated agnostic of their identity. During \textit{instance identification}, an exploration algorithm is deployed to interact with each object. The contact data resulting from the interaction with each object is matched against the contact data of the target object for identification. Lastly, the position of the identified object is used to \textit{grasp} and retrieve it. In each step, both interaction and inference algorithms are designed to minimize object motion and to integrate sparse tactile information for dealing with the challenges mentioned in Section~\ref{sec:introduction}. We train the identification model in simulation and transfer it to the real world, whereas the localization and interaction strategies are directly implemented in the real world. 

\subsection{Object Localization with Tactile Sensing}
\label{sec:localization}
For the task of interest, an object localization 
The main challenge in object localization is that object motion while localizing the $N$-th object may disrupt the estimates for any of the $N-1$ objects localized previously. Minimizing object motion is further complicated when object masses, shapes, and friction properties are unknown. Assuming that no two objects are vertically stacked and that objects do not touch each other, our approach to object localization consists of two steps: first, the workspace is discretized into a square grid ($H$) with fixed side length $\delta = 5cm$ (roughly half the average width of the target objects) and every location on the grid is probed to estimate if it is occupied by the object. The result is a binary \textit{object occupancy map} (i.e., $H[i,j] \leftarrow \{0,1\}$). Next, $H$ is clustered to obtain approximate locations for each object.

\begin{figure}[!tb]
    \centering
    \includegraphics[width=0.7\linewidth]{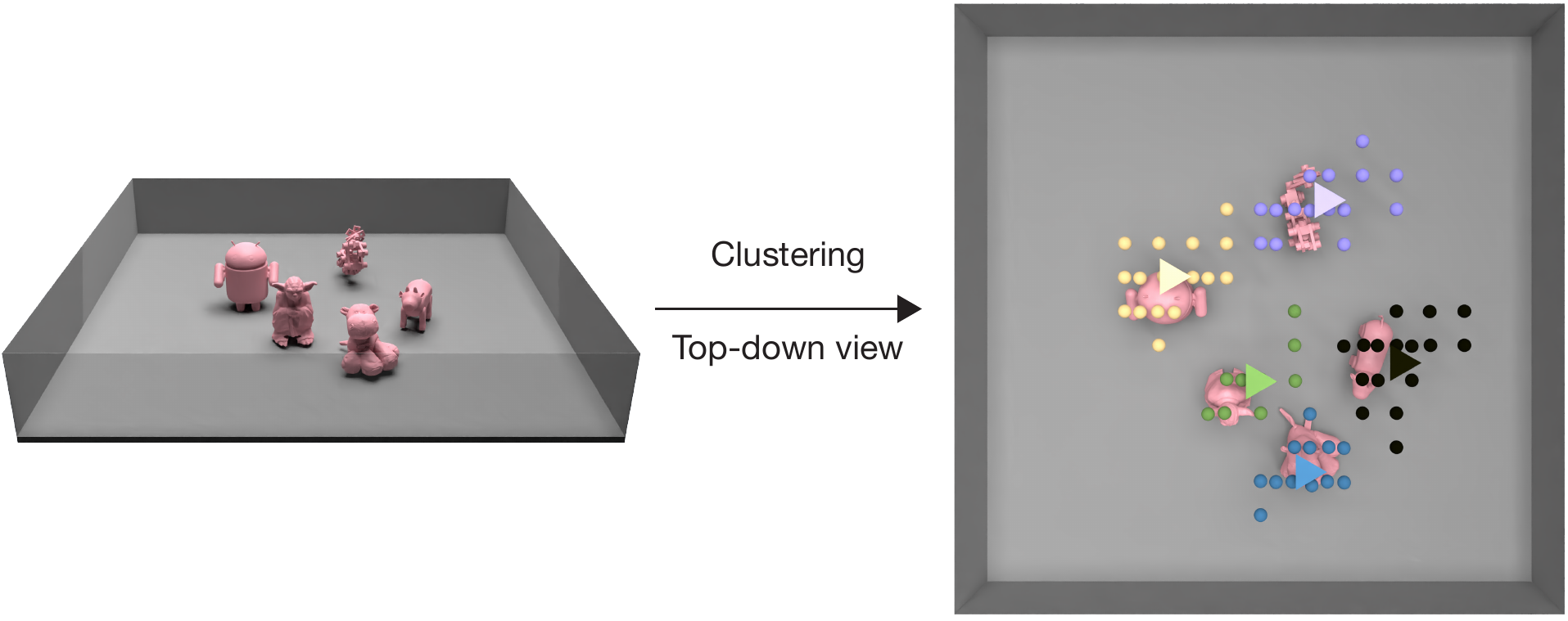}
    \caption{\footnotesize{Visualization of planar occupancy and resulting clusters (the dots) during localization compared with true object centers (the triangles). While showing non-zero error, the relative object centers are well predicted enough for subsequent object identification.}}
    \label{fig:localization_viz}
    \vspace{-0.7cm}
\end{figure}

To obtain the occupancy map, $H[i,j]$ we move the gripper to the center of the grid location $(i,j)$ and at a pre-specified height above the bottom of the bin/table. The gripper is commanded to move down along the normal to the plane (Fig~\ref{fig:pipeline}). To minimize object motion, as soon as any contact is encountered or if the tip of the gripper reaches the surface of the table, the arm's motion is stopped. If contact was encountered, the grid position is deemed to be occupied (i.e., $H[i,j] = 1$). The procedure is repeated for all grid locations. The occupancy map is clustered using K-means algorithm~\cite{lloyd82kmeans}, with a known value of $K$ corresponding to the number of objects in the bin. This procedure yields a list of approximate object center positions $\left[o_c^1, o_c^2, \dots, o_c^N\right]$ (Fig~\ref{fig:localization_viz}), which can then be utilized for fine-grained object identification. 

\subsection{Dynamic Object Identification}
\label{sec:identification} 
The localization process above provides an occupancy map and object locations, which is insufficient to estimate the 3D shape of objects required to perform identification. Prior works have found that random
One naive strategy to get more interaction data is to randomly move the fingers, and once in a while, the fingers will touch the object. However, such a naive strategy makes it hard to identify objects. The reason is that the object might move a lot due to random finger motions, causing noise in the data. When the object does not move, the fingertip locations when fingers make contact with the object represent the points on the object's surface. However, if the object moves a lot, the touch points are effects of both the object's geometry as well as the object's motion, which makes it more challenging to identify what object it is. Therefore, it is beneficial to have an interaction policy that explores the object's surface but also tries to minimize the object's motion.

After collecting such data, the next step is identifying whether the object is the target object we are searching for. A straightforward way is to train a classifier that outputs the object ID and see if it matches the target object ID. However, such a classifier cannot generalize to novel objects. Therefore, instead, we use contrastive learning to learn an embedding space in which we can determine whether the object is the same as the target object based on the distance between their embedding vectors. 

\begin{figure}[!h]
    \centering
    \includegraphics[width=1.\linewidth]{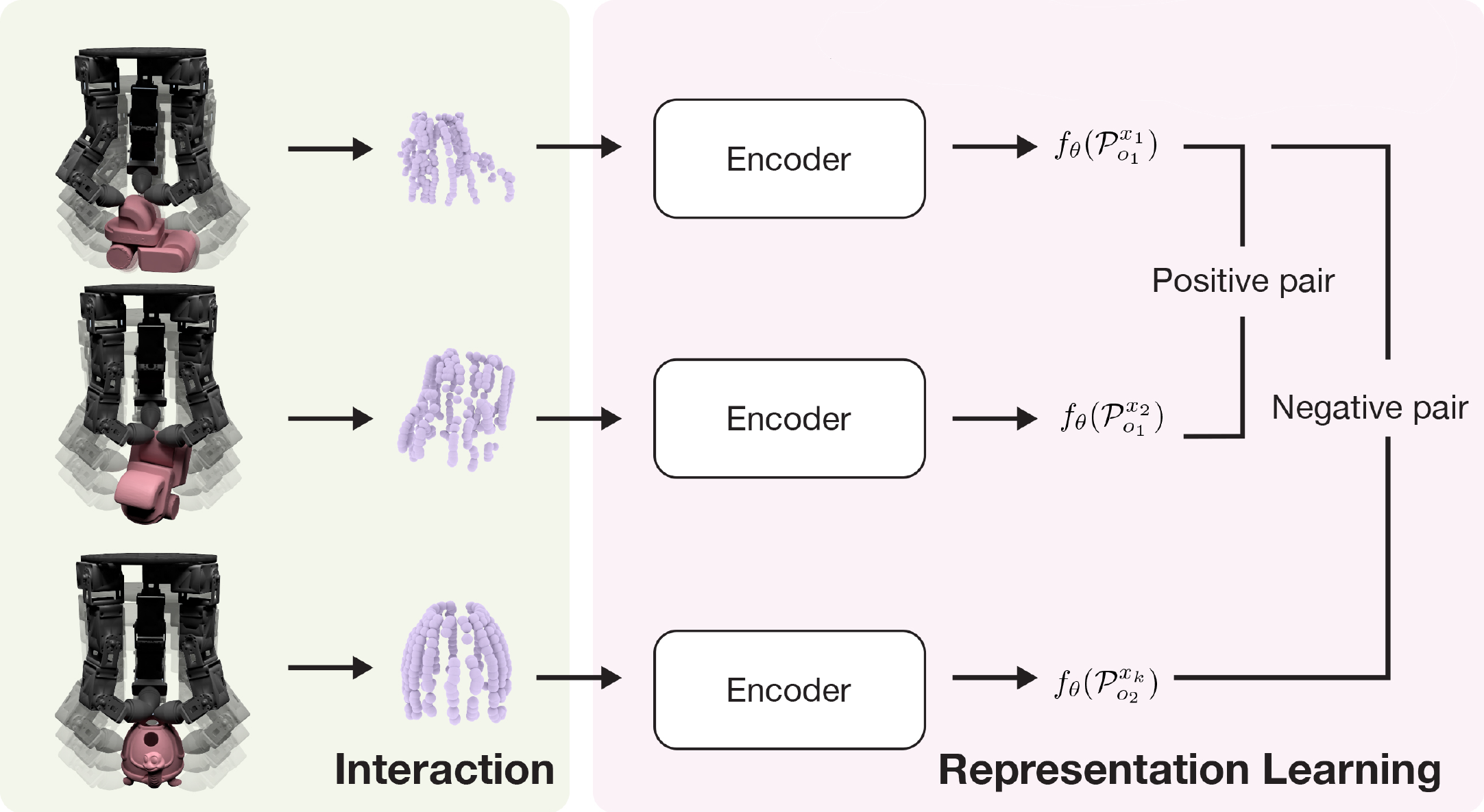}
    \caption{\footnotesize{Depiction of the contrastive representation learning architecture for object identification. The sequence of contact points goes through a transformer based encoder to provide embeddings that are trained with the InfoNCE loss. Affinity in this representation space can then be used for object identification}} 
    \label{fig:representation_learning}
\end{figure}

\paragraph{Collecting Interaction Data} 
Instead of having the hand doing unstructured exploration around the object, which obtains very few useful interaction data and can cause the object to have a big motion, we devised a radial sliding strategy. It tries to move the fingers to follow the contours of the objects in the vertical direction. We hypothesize that such slices of object contours along the vertical axis would be sufficient for identifying an object (as described in the next section) since they give a notion of object shape. However, in our experiments, we found that having the fingertips move along the object contours when object geometry is complex tends to cause the object to move. Instead, we command the fingers to \textit{tap} along the object contours (\textit{radial tapping strategy}). Tapping significantly reduces the object motion and allows us to identify free objects much more effectively than sliding. This is because the number of points of contact is reduced and the force inward can be controlled more carefully. As shown in Fig \ref{fig:representation_learning}, we first reset the fingers to their initial positions, and then close the fingers (move the fingertips inwards to touch the objects), when touch happens on any of the fingertips ($c^t_j=1$), we record the fingertip position $p^t_j$ and stop moving the finger. Since each touch only generates one data point (the fingertip position $p_j^t$), we get up to three points at each time we close the fingers (potentially one from each finger). We move the fingers upward by a small distance and repeat such a process until the fingers touch the top of the object. By doing so, we get a sequence of contact positions $\mathcal{P}_{o}^x = \{p_0^0, p_1^0, p_2^0, p_0^1, p_1^1, p_2^1, \dots, p_0^N, p_1^N, p_2^N\}$ for a series of $N$ taps for object $o$ at an unknown pose $x$, where each contact point $p$ is a point in $\mathbb{R}^3$. 

Some things to note --- Firstly, the actual contact forces do not need to be used since we are just detecting binary contact on the surface. Secondly, the object itself may move since the fingers are not \emph{guaranteed} to perfectly cage the object while the contact is being made. To account for potential object motion, we perform object relocalization by estimating the direction of motion from the contact points at each time step. Specifically, at every point in time, we maintain a current estimate $\hat{c}$ for the current object center. Each radial tap gives us a sequence of contact points $\{p_i\}$, which we average to get a single point $p$. We then perform an exponential smoothing update $\hat{c}_{new} = \gamma \hat{c}_{old} + (1-\gamma) p$ to get a smoothed new estimate, and move the hand in the $xy$-plane to $\hat{c}_{new}$. This allows us to correct for local object motions that may occur when interacting with the object without knowing the object model beforehand. Note that our goal is not to accurately estimate the center of the object. A rough estimate is sufficient for the hand to track its location. As long as the three fingers can surround the object, we can collect meaningful interaction data.

\paragraph{Representation Learning for Object Identification}
The contact information we get from the interaction policy is a sequence of contact positions $\mathcal{P}_{o}^x$, which can be viewed as a very sparse point cloud. Our goal is to identify whether the obtained point cloud $\mathcal{P}_{o}^{x_1}$ and the point cloud obtained by interacting with the target object $\mathcal{P}_{o_{\text{tgt}}}^{x_2}$ belong to the same object. Note that even if they are the same object, the object can be in different poses. Therefore, we need a shape representation that is agnostic to the object orientation but still allows distinguishing between different objects. We can learn such representation using contrastive learning~\cite{Hadsell2006siamese, oord2018cpc, zhang21contrastivetactile, chen20simclr} --- represent touches on the same object at different poses similarly and touches across different objects differently. One potential representation learning objective is the InfoNCE loss (as also discussed in ~\cite{oord2018cpc}). Assuming we have some parametric encoder $f$ with parameters $\theta$, this loss can be expressed as:\[\mathcal{L}_{\text{NCE}}(\theta) = \mathbb{E}_{(x, x^+) \sim \mathcal{D}_{o}, \{x_k\}\sim\mathcal{D}}\Biggl[ \log\frac{\exp{f_\theta(x)\cdot f_\theta(x^+)}}{\sum \exp{f_\theta(x)\cdot f_\theta(x_k)}} \Biggr]\] where the positive distribution $\mathcal{D}_{o}$ consists of touch sequences sampled from different poses of the same object $o$ and the negative distribution $\mathcal{D}$ is touch sequences sampled from arbitrary poses of different objects. We treat the touch data (sequences of contact positions $\mathcal{P}_{o}^x$) as a sequential time serie, and parameterize the encoder $f_\theta$ with a transformer architecture.

Once the representation is learned, the object of interest can be identified by comparing the cosine similarity of touch sequences in the representation space $z = f_\theta(\mathcal{P}_{o}^{x})$ of the touches made blindly across various objects in the bin with the touches made on the target object outside of the bin. This identified object can be used for object retrieval using a grasping controller.

While more complex schemes could be developed for grasping ~\cite{chen2022system, duan21dexterous}, we found that the hand lends itself very naturally to grasping with a simple caging policy. Concretely, we simply close the fingers until the fingertips touch the object with some amount of force. The touch sensors on the fingertips are used to tell whether the fingertips touch the object with a sufficient amount of contact area. After all three fingers touch the object with a sufficient amount of contact, we lift the arm to grasp the object. We depict this policy in Fig~\ref{fig:pipeline}, and deploy it in an object-agnostic manner to grasp objects which have been localized and identified. 

\section{Experimental Evaluation}
In this section, we aim to answer the following questions: (1) How effective is the overall pipeline at identifying and grasping particular object instances from tactile sensing alone? (2) Can a dexterous hand with tactile sensing localize movable objects without visual input? (3) Is the data obtained by interacting with objects using the radial tapping strategy described in Section~\ref{sec:identification} effective for classification? (4) Is the radial tapping interaction strategy for object identification effective for movable objects? (5) Is representation learning using the scheme in Section~\ref{sec:identification} effective for identifying novel objects?

We used the 150 object meshes collected in \cite{chen2022visual}. The meshes are from various datasets including Google Scanned Objects~\cite{downs2022google} and ShapeNet~\cite{shapenet2015}. We built the simulation environment in PyBullet. The 3-D printed versions of the objects were used for testing in the real world. For evaluation, we create test scenes by randomly choosing $K$ objects from this set and placing them in random poses in a bin.

\subsection{Baselines and Evaluation Metrics}
While no entire system exists that completes the entire task described in this work, we compare it with several prior works that perform localization, identification, or grasping. For localization, we compare with ~\cite{murali18blind} which uses particle filters for localization. To understand the importance of our particular interaction strategy, sensing modality and other design choices in terms of being able to collect information without moving the object significantly during object identification, we compared with two baselines: 1) radial tapping with noisy contact detection rather than accurate binary contact direction (noisy contact) to test the importance of accurately binarizing contact, 2) radial tapping without hand relocalization in cases where the object slips as discussed in Section~\ref{sec:identification} (no relocalization) to show the importance of the relocalization strategy. To understand the impact of specific model architecture, we compared with two baselines: 1) an LSTM~\cite{bottcher2021object} rather than a transformer with the InfoNCE objective, 2) using Triplet loss~\cite{schroff2015facenet, chopra2005learning} instead of InfoNCE loss function, to understand the importance of the particular choice of transformer and infoNCE objectives.

\noindent \textbf{Evaluation Metrics}: For localization, we measure the error (in cm) between the identified object center and the ground truth object center of mass (clustering error). Additionally, we measure displacement from the original object positions to determine how disruptive exploration is (perturbation error). We evaluated this for scenes with $3$ objects. For identification, we measure the success percentage in identifying the correct object out of $5$ unique object instances that are randomly placed in the bin, as well as the number of successful taps made with the object. 

\begin{figure}[!t]
    \centering
    \includegraphics[width=1.\linewidth]{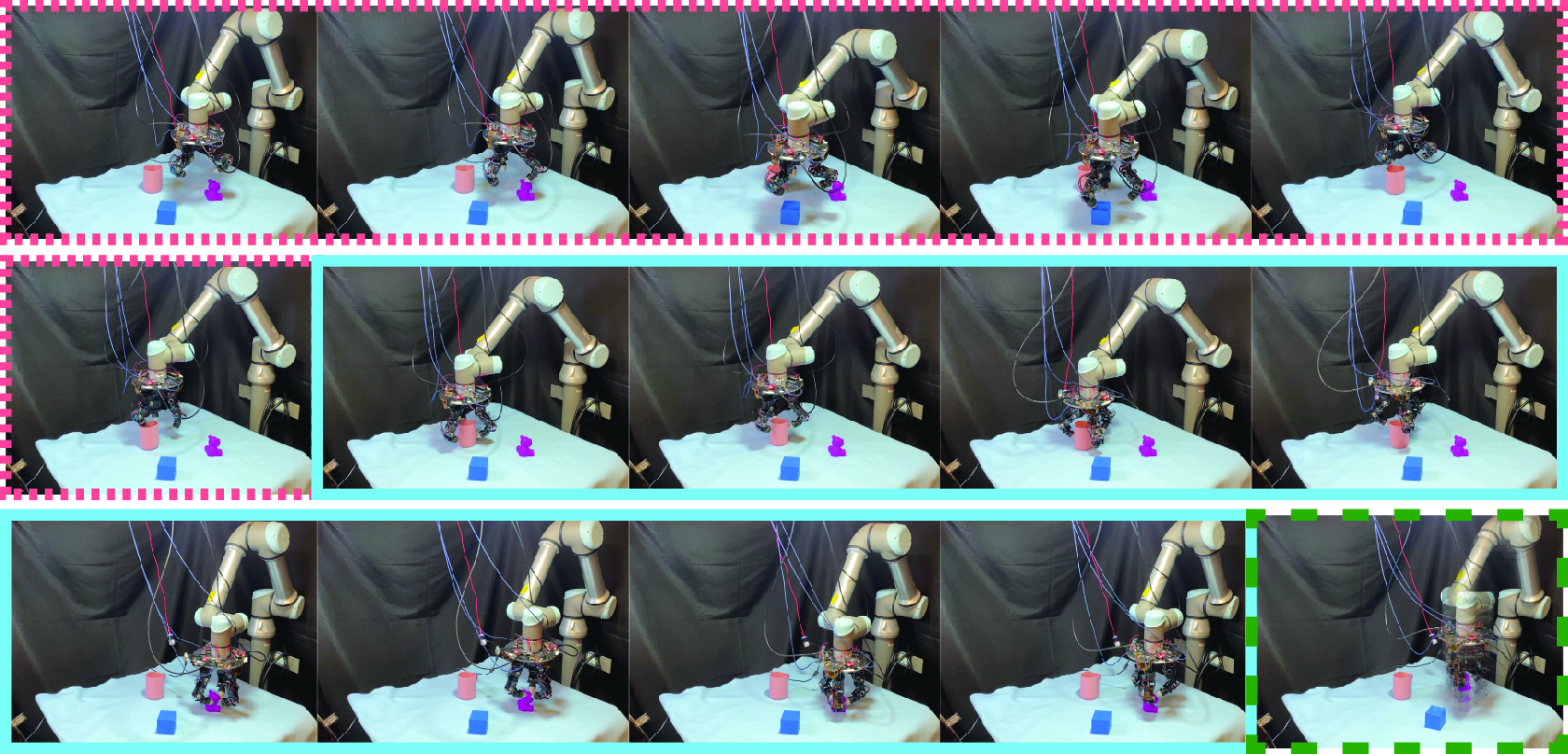}
    \caption{\footnotesize{Film strip depicting phases of real world localization and identification with our pipeline. We see that objects move minimally, while shapes are successfully identified and grasped. Red box shows the localization, blue box shows the interaction, and the green box shows the grasping.}}
    \label{fig:real_execution}
    \vspace{-0.5cm}
\end{figure}

\subsection{Simulation Results}
\label{sec:sim_results}
We report results on localizing objects in different scenes in simulation with various different objects in Table~\ref{table:localization}. We see that our proposed clustering scheme gets within $3.6$ cm on average across various objects while ensuring minimal displacement of only $1.7$ cm during exploration when evaluated on test objects approximately 10 to 15 cm in diameter. In comparison, the particle filtering scheme used in ~\cite{murali18blind} performs competitively with ours on scenes with single objects, achieving an average of $4.5cm$ of error. However, when increasing the number of objects to three, the particle filtering struggles and the error increases to $6.4cm$ on average. This is likely due to the fact that increasing the number of objects also increases the dimensionality of the particles, meaning exponentially more particles are needed to achieve more accuracy.

\begin{table}[!h]
\caption{\footnotesize{Results on object localization in simulation. For both strategies, we measure the success rate (defined as when each predicted center is within $\delta = 7.5cm$ of the object center), the average prediction error, and the average displacement of the objects during the localization. Our clustering strategy scales to multiple objects much better than particle filtering, likely because multiple objects introduce a higher dimensional search space.}}
\centering
\resizebox{\columnwidth}{!}{
\begin{tabular}{cccc} 
\hline
Method          & Success Rate & Center Error & Object Displacement  \\ 
\hline
Ours (1 object)           & $99.2\%\pm 0.2\%$       & $3.2\pm 0.1$cm        & $1.8\pm 0.1$cm                \\
Particle Filter (1 object) ~\cite{murali18blind} & $94.6\%\pm 0.6\%$       & $4.5\pm 0.3$cm        & $1.5\pm 0.1$cm                \\
Ours (3 objects)           & $91.3\%\pm 1.7\%$       & $3.6\pm 0.2$cm        & $1.7\pm 0.1$cm\\
Particle Filter (3 objects) ~\cite{murali18blind} & $52.8\%\pm 1.4\%$       & $6.4\pm 0.1$cm        & $2.4\pm 0.1$cm                \\
\hline
\end{tabular}}

\label{table:localization}
\end{table}

Next, we show in Table~\ref{table:interaction}, that our proposed identification technique is able to select the correct object out of five novel objects with $69.8\%$ accuracy.

\textbf{Interaction:} To understand why the interaction strategy is better than alternatives, we study the impact of removing individual elements of the interaction strategy. Table \ref{table:interaction} shows the results of these ablations. First, we add random noise into the sensors, compromising the system's ability to quickly detect contact. This results in identification success dropping from $69.8\%$ to $54.4\%$, likely due to greatly increased object motion. We also investigate the effects of removing the relocalization policy described in Section~\ref{sec:identification}. Without the ability to adapt to object motion, identification accuracy drops to $58.5\%$, a drop of over $10\%$. Therefore, we can conclude that both our relocalization policy, as well as accurate contact detection, are necessary to achieve our system's accuracy. 

\begin{table}[!tb]
\centering
\resizebox{\columnwidth}{!}{
\begin{tabular}{cccc} 
\hline
Method            & Accuracy (Validation Set) & Accuracy (Training Set) &  Successful Taps  \\ 
\hline
Ours              & $\mathbf{69.8\%\pm 1.2\%}$ &  $88.4\%\pm 0.9\%$   & $\mathbf{228.4\pm 0.7}$             \\
No Relocalization & $58.5\%\pm 1.4\%$ & $61.8\%\pm 1.3\%$ & $157.1\pm 0.8$                        \\
Noisy Sensors     & $54.4\%\pm 1.4\%$ &  $59.3\%\pm 1.4\%$           & $183.8\pm 0.7$                        \\
\hline
\end{tabular}}
\caption{\footnotesize{Effectiveness of interaction strategies on object identification in simulation. We measure for each policy the overall accuracy when selecting out of five (previously unseen) objects on both the training and validation object sets, as well as the average number of successful taps the policy gets from the object. These experiments show that both our object position estimation as well as accurate contact detection are essential for successful object identification.}}
\label{table:interaction}
\end{table}

\textbf{Identification:} From Table~\ref{table:arch_comparison}, we see that the transformer with the InfoNCE objective achieves a higher success rate than the alternatives. In particular, training with InfoNCE loss improves performance over using Triplet loss 

Finally, in simulation, we evaluate the combined pipeline of object localization and identification (leaving grasping for real-world evaluation in Section~\ref{sec:real_results}) in terms of the percentage of trials where the successful object is identified and localized. We find in Table~\ref{table:full_pipeline} that our pipeline is able to accomplish a success rate of $76.8\%$ in simulation.

\begin{table}[!tb]
\centering
\resizebox{\columnwidth}{!}{
\begin{tabular}{ccc} 
\hline
Method                    & Accuracy (5 objects) & Accuracy (3 objects)  \\ 
\hline
Transformer + InfoNCE loss (Ours)                      & $69.8\%\pm 1.2\%$  & $80.3\%\pm 1.0\%$    \\
LSTM + InfoNCE loss              & $65.7\%\pm 1.3\%$   & $77.0\%\pm 1.2\%$                      \\
Transformer + Triplet loss    & $52.3\%\pm 1.4\%$   & $64.8\%\pm 1.3\%$                      \\
\hline
\end{tabular}}
\caption{\footnotesize{Effects of architecture and objective on object identification in simulation. Using a more expressive architecture like a transformer helps with classification, while the InfoNCE loss outperforms using a triplet loss.}}
\label{table:arch_comparison}
\vspace{-0.3cm}
\end{table}

\subsection{Real World Results}
\label{sec:real_results}

\begin{table}[!h]
\centering
\resizebox{\columnwidth}{!}{
\begin{tabular}{cccc} 
\hline
Success Rate               & Overall &  Localization & Identification\\ 
\hline
Sim            & $76.8\%\pm 2.3\%$       & $91.3\%\pm 1.7\%$    & $80.3\%\pm 1.0\%$                                                          \\
Real   &         $59.4\%\pm 11.3\%$     &   $81.0\%\pm 9.0\%$       &  $75.7\%\pm 9.8\%$                                                              \\
\hline
\end{tabular}}
\caption{\footnotesize{Results on full pipeline object retrieval in simulation and the real world. We find that there are expected drops in performance from simulation to the real system, but the identification is still able to significantly outperform random chance}}
\label{table:full_pipeline}
\vspace{-0.3cm}
\end{table}

The pipeline transfers relatively well to the real world. Due to the lack of ground truth object positions running on the real system, we qualitatively evaluate pipeline stage success. Specifically, we deem localization to be successful if the robot is able to get its three fingers around each object, and grasping to be successful if the robot is able to lift the chosen object. When evaluated on three objects at a time, we find that localization has a success rate of $78\%$. The object identification model trained in simulation has a success rate of $73\%$ in the real world. And similarly, grasping is able to accomplish a success rate of $71\%$. When this entire system is executed sequentially, it accomplishes a success rate of $59.4\%$, as shown in Fig~\ref{fig:real_execution}. While this leaves room for improvement, in each part of the pipeline, it is significantly better than chance and much better than a random policy.

\subsection{Ablations}
\label{sec:ablations}

To understand how much moving objects affect classification performance, we perform an ablation study in Table~\ref{table:interaction_static} repeating the comparisons in Section~\ref{sec:sim_results} with a fixed stationary object. We find that classification is significantly easier, showing the difficulty of scenarios with moving objects.

\begin{table}[!tb]
\centering
\resizebox{\columnwidth}{!}{
\begin{tabular}{ccc} 
\hline
Method & Accuracy (5 objects) & No. of Successful Taps  \\ 
\hline
Ours (Moving Objects)   & $69.8\%\pm 1.2\%$     & $228.4\pm 0.7$             \\
Ours (Static Objects)   & $86.2\%\pm 0.9\%$     & $235.4\pm 1.2$             \\
\hline
\end{tabular}}
\caption{\footnotesize{Effectiveness of interaction strategies on object identification in simulation for \emph{static} objects. We run our object interaction strategy on both moving and static objects and find that static objects are easier to classify than moving ones, showing how challenging our problem setting is}}
\label{table:interaction_static}
\end{table}

\begin{table}[!h]
\centering
\resizebox{\columnwidth}{!}{
\begin{tabular}{cccc} 
\hline
Friction Coefficient          & $\mu = 0.5$     & $\mu = 0.25$    & $\mu = 0.1$\\ 
\hline
Localization Accuracy  & $91.3\%\pm 1.7\%$ & $82.6\%\pm 1.8\%$ & $17.3\%\pm 1.7\%$    \\
Identification & $69.8\%\pm 1.2\%$          &   $65.3\%\pm 1.3\%$         & $53.8\%\pm1.4\%$\\
\hline
\end{tabular}}
\caption{\footnotesize{Effect of the coefficient of friction ($\mu$) on localization and classification performance in simulation. As friction reduce, localization becomes much more difficult and identification accuracy also decreases due to increased object movement.}}
\label{table:physicalparams}
\vspace{-0.3cm}
\end{table}

To understand just how much these methods help with moving objects, we also ran an ablation study where we performed comparisons in Table~\ref{table:physicalparams} as we change object friction and mass. The localization performance degrades as we use small friction and mass values, especially low friction. This is expected as on a slippery surface, an object moves more easily and suffers from large motion even if the robot hand applies a small force on it. However, our identification pipeline suffers a much smaller performance drop, indicating the capability of our method in handling non-static objects. 

\section{Discussion}
\label{sec:dic}

The pipeline proposed in this work is only a starting point for tactile-only object localization and retrieval. While we have designed a strategy using tapping to minimize object movement, an interesting future research direction would be to explore how to localize, identify and grasp objects that can move substantially.

\section{Author Contributions and Acknowledgements}
This work is supported by grants from the Toyota Research Institute, DARPA Machine Common Sense program, ONR MURI grant N00014-22-1-2740, and the NSF Graduate Research Fellowship.

\textbf{Sameer Pai}: Led the system design, simulation/real-world platform development, experiment running, and paper writing. \textbf{Tao Chen}: Contributed substantially to system design, simulation/real-world platform development, paper writing, and advised Sameer. \textbf{Megha Tippur}: Developed and built the tactile sensors. \textbf{Edward Adelson}: Advised and supported the tactile sensor development. \textbf{Abhishek Gupta}: Oversaw and advised the project. Also contributed significantly to simulation environment building and paper writing. \textbf{Pulkit Agrawal}: Conceived, oversaw, and advised the project and contributed to writing.

\bibliographystyle{unsrt}  
\bibliography{paper}

\begin{thebibliography}{10}

\bibitem{bottcher2021object}
Wolfgang Bottcher, Pedro Machado, Nikesh Lama, and Thomas~M McGinnity.
\newblock Object recognition for robotics from tactile time series data
  utilising different neural network architectures.
\newblock In {\em 2021 International Joint Conference on Neural Networks
  (IJCNN)}, pages 1--8. IEEE, 2021.

\bibitem{bauza2019tactile}
Maria Bauza, Oleguer Canal, and Alberto Rodriguez.
\newblock Tactile mapping and localization from high-resolution tactile
  imprints.
\newblock In {\em 2019 International Conference on Robotics and Automation
  (ICRA)}, pages 3811--3817. IEEE, 2019.

\bibitem{chopra05contrastive}
Sumit Chopra, Raia Hadsell, and Yann LeCun.
\newblock Learning a similarity metric discriminatively, with application to
  face verification.
\newblock In {\em 2005 {IEEE} Computer Society Conference on Computer Vision
  and Pattern Recognition {(CVPR} 2005), 20-26 June 2005, San Diego, CA,
  {USA}}, pages 539--546. {IEEE} Computer Society, 2005.

\bibitem{oord2018cpc}
A{\"{a}}ron van~den Oord, Yazhe Li, and Oriol Vinyals.
\newblock Representation learning with contrastive predictive coding.
\newblock {\em CoRR}, abs/1807.03748, 2018.

\bibitem{kappassov2015tactile}
Zhanat Kappassov, Juan-Antonio Corrales, and V{\'e}ronique Perdereau.
\newblock Tactile sensing in dexterous robot hands.
\newblock {\em Robotics and Autonomous Systems}, 74:195--220, 2015.

\bibitem{sundaram2019learning}
Subramanian Sundaram, Petr Kellnhofer, Yunzhu Li, Jun-Yan Zhu, Antonio
  Torralba, and Wojciech Matusik.
\newblock Learning the signatures of the human grasp using a scalable tactile
  glove.
\newblock {\em Nature}, 569(7758):698--702, 2019.

\bibitem{boutry2018hierarchically}
Clementine~M Boutry, Marc Negre, Mikael Jorda, Orestis Vardoulis, Alex Chortos,
  Oussama Khatib, and Zhenan Bao.
\newblock A hierarchically patterned, bioinspired e-skin able to detect the
  direction of applied pressure for robotics.
\newblock {\em Science Robotics}, 3(24):eaau6914, 2018.

\bibitem{cutkosky2008force}
Mark~R Cutkosky, Robert~D Howe, and William~R Provancher.
\newblock Force and tactile sensors.
\newblock {\em Springer Handbook of Robotics}, 100:455--476, 2008.

\bibitem{bhirangi2021reskin}
Raunaq Bhirangi, Tess Hellebrekers, Carmel Majidi, and Abhinav Gupta.
\newblock Reskin: versatile, replaceable, lasting tactile skins.
\newblock {\em arXiv preprint arXiv:2111.00071}, 2021.

\bibitem{ward2018tactip}
Benjamin Ward-Cherrier, Nicholas Pestell, Luke Cramphorn, Benjamin Winstone,
  Maria~Elena Giannaccini, Jonathan Rossiter, and Nathan~F Lepora.
\newblock The tactip family: Soft optical tactile sensors with 3d-printed
  biomimetic morphologies.
\newblock {\em Soft robotics}, 5(2):216--227, 2018.

\bibitem{kuppuswamy2020soft}
Naveen Kuppuswamy, Alex Alspach, Avinash Uttamchandani, Sam Creasey, Takuya
  Ikeda, and Russ Tedrake.
\newblock Soft-bubble grippers for robust and perceptive manipulation.
\newblock In {\em 2020 IEEE/RSJ International Conference on Intelligent Robots
  and Systems (IROS)}, pages 9917--9924. IEEE, 2020.

\bibitem{lambeta2020digit}
Mike Lambeta, Po-Wei Chou, Stephen Tian, Brian Yang, Benjamin Maloon,
  Victoria~Rose Most, Dave Stroud, Raymond Santos, Ahmad Byagowi, Gregg
  Kammerer, et~al.
\newblock Digit: A novel design for a low-cost compact high-resolution tactile
  sensor with application to in-hand manipulation.
\newblock {\em IEEE Robotics and Automation Letters}, 5(3):3838--3845, 2020.

\bibitem{she2021cable}
Yu~She, Shaoxiong Wang, Siyuan Dong, Neha Sunil, Alberto Rodriguez, and Edward
  Adelson.
\newblock Cable manipulation with a tactile-reactive gripper.
\newblock {\em The International Journal of Robotics Research},
  40(12-14):1385--1401, 2021.

\bibitem{wang2020swingbot}
Chen Wang, Shaoxiong Wang, Branden Romero, Filipe Veiga, and Edward Adelson.
\newblock Swingbot: Learning physical features from in-hand tactile exploration
  for dynamic swing-up manipulation.
\newblock In {\em 2020 IEEE/RSJ International Conference on Intelligent Robots
  and Systems (IROS)}, pages 5633--5640. IEEE, 2020.

\bibitem{dong2021tactile}
Siyuan Dong, Devesh~K Jha, Diego Romeres, Sangwoon Kim, Daniel Nikovski, and
  Alberto Rodriguez.
\newblock Tactile-rl for insertion: Generalization to objects of unknown
  geometry.
\newblock In {\em 2021 IEEE International Conference on Robotics and Automation
  (ICRA)}, pages 6437--6443. IEEE, 2021.

\bibitem{chaudhury2022using}
Arkadeep~Narayan Chaudhury, Timothy Man, Wenzhen Yuan, and Christopher~G
  Atkeson.
\newblock Using collocated vision and tactile sensors for visual servoing and
  localization.
\newblock {\em IEEE Robotics and Automation Letters}, 7(2):3427--3434, 2022.

\bibitem{yang2015object}
Jingwei Yang, Huaping Liu, Fuchun Sun, and Meng Gao.
\newblock Object recognition using tactile and image information.
\newblock In {\em 2015 IEEE International Conference on Robotics and
  Biomimetics (ROBIO)}, pages 1746--1751. IEEE, 2015.

\bibitem{corradi2017object}
Tadeo Corradi, Peter Hall, and Pejman Iravani.
\newblock Object recognition combining vision and touch.
\newblock {\em Robotics and biomimetics}, 4(1):1--10, 2017.

\bibitem{yuan2018active}
Wenzhen Yuan, Yuchen Mo, Shaoxiong Wang, and Edward~H Adelson.
\newblock Active clothing material perception using tactile sensing and deep
  learning.
\newblock In {\em 2018 IEEE International Conference on Robotics and Automation
  (ICRA)}, pages 4842--4849. IEEE, 2018.

\bibitem{li2013sensing}
Rui Li and Edward~H Adelson.
\newblock Sensing and recognizing surface textures using a gelsight sensor.
\newblock In {\em Proceedings of the IEEE Conference on Computer Vision and
  Pattern Recognition}, pages 1241--1247, 2013.

\bibitem{pezzementi2011mapping}
Zachary~A. Pezzementi, Caitlin Reyda, and Gregory~D. Hager.
\newblock Object mapping, recognition, and localization from tactile geometry.
\newblock In {\em {IEEE} International Conference on Robotics and Automation,
  {ICRA} 2011, Shanghai, China, 9-13 May 2011}, pages 5942--5948. {IEEE}, 2011.

\bibitem{pohtongkam2021tactile}
Somchai Pohtongkam and Jakkree Srinonchat.
\newblock Tactile object recognition for humanoid robots using new designed
  piezoresistive tactile sensor and dcnn.
\newblock {\em Sensors}, 21(18):6024, 2021.

\bibitem{lin2019multimodal}
Justin Lin, Roberto Calandra, and Sergey Levine.
\newblock Learning to identify object instances by touch: Tactile recognition
  via multimodal matching.
\newblock {\em CoRR}, abs/1903.03591, 2019.

\bibitem{jamali2016activetactile}
Nawid Jamali, Carlo Ciliberto, Lorenzo Rosasco, and Lorenzo Natale.
\newblock Active perception: Building objects' models using tactile
  exploration.
\newblock In {\em 16th {IEEE-RAS} International Conference on Humanoid Robots,
  Humanoids 2016, Cancun, Mexico, November 15-17, 2016}, pages 179--185.
  {IEEE}, 2016.

\bibitem{uriel2013contour}
Uriel Martinez{-}Hernandez, Giorgio Metta, Tony~J. Dodd, Tony~J. Prescott,
  Lorenzo Natale, and Nathan~F. Lepora.
\newblock Active contour following to explore object shape with robot touch.
\newblock In {\em 2013 World Haptics Conference, {WHC} 2013, Daejeon, Korea
  (South), April 14-17, 2013}, pages 341--346. {IEEE}, 2013.

\bibitem{yi2016activeexploration}
Zhengkun Yi, Roberto Calandra, Filipe Veiga, Herke van Hoof, Tucker Hermans,
  Yilei Zhang, and Jan Peters.
\newblock Active tactile object exploration with gaussian processes.
\newblock In {\em 2016 {IEEE/RSJ} International Conference on Intelligent
  Robots and Systems, {IROS} 2016, Daejeon, South Korea, October 9-14, 2016},
  pages 4925--4930. {IEEE}, 2016.

\bibitem{driess2017activetactile}
Danny Drie{\ss}, Peter Englert, and Marc Toussaint.
\newblock Active learning with query paths for tactile object shape
  exploration.
\newblock In {\em 2017 {IEEE/RSJ} International Conference on Intelligent
  Robots and Systems, {IROS} 2017, Vancouver, BC, Canada, September 24-28,
  2017}, pages 65--72. {IEEE}, 2017.

\bibitem{xu2022tandem}
Jingxi Xu, Shuran Song, and Matei Ciocarlie.
\newblock Tandem: Learning joint exploration and decision making with tactile
  sensors.
\newblock {\em arXiv preprint arXiv:2203.00798}, 2022.

\bibitem{Karayiannidis2014OnlineCP}
Yiannis Karayiannidis, Christian Smith, Francisco~E. Vina, and Danica Kragic.
\newblock Online contact point estimation for uncalibrated tool use.
\newblock {\em 2014 IEEE International Conference on Robotics and Automation
  (ICRA)}, pages 2488--2494, 2014.

\bibitem{likar2014external}
Nejc Likar and Leon {\v{Z}}lajpah.
\newblock External joint torque-based estimation of contact information.
\newblock {\em International Journal of Advanced Robotic Systems}, 11(7):107,
  2014.

\bibitem{li2014localization}
Rui Li, Robert Platt, Wenzhen Yuan, Andreas ten Pas, Nathan Roscup, Mandayam~A
  Srinivasan, and Edward Adelson.
\newblock Localization and manipulation of small parts using gelsight tactile
  sensing.
\newblock In {\em 2014 IEEE/RSJ International Conference on Intelligent Robots
  and Systems}, pages 3988--3993. IEEE, 2014.

\bibitem{luo2017visualmap}
Shan Luo, Wenxuan Mou, Kaspar Althoefer, and Hongbin Liu.
\newblock Localizing the object contact through matching tactile features with
  visual map.
\newblock {\em CoRR}, abs/1708.04441, 2017.

\bibitem{koval2015manifold}
Michael~C. Koval, Nancy~S. Pollard, and Siddhartha~S. Srinivasa.
\newblock Pose estimation for planar contact manipulation with manifold
  particle filters.
\newblock {\em Int. J. Robotics Res.}, 34(7):922--945, 2015.

\bibitem{kissoum2021simultaneous}
Ghani Kissoum and Veronique Perdereau.
\newblock Simultaneous tactile localization and reconstruction of an object
  during robotic manipulation.
\newblock In {\em 2021 20th International Conference on Advanced Robotics
  (ICAR)}, pages 948--954. IEEE, 2021.

\bibitem{murali18blind}
Adithyavairavan Murali, Yin Li, Dhiraj Gandhi, and Abhinav Gupta.
\newblock Learning to grasp without seeing.
\newblock In Jing Xiao, Torsten Kr{\"{o}}ger, and Oussama Khatib, editors, {\em
  Proceedings of the 2018 International Symposium on Experimental Robotics,
  {ISER} 2018, Buenos Aires, Argentina, November 5-8, 2018}, volume~11 of {\em
  Springer Proceedings in Advanced Robotics}, pages 375--386. Springer, 2018.

\bibitem{petrovskaya11localization}
Anna Petrovskaya and Oussama Khatib.
\newblock Global localization of objects via touch.
\newblock {\em {IEEE} Trans. Robotics}, 27(3):569--585, 2011.

\bibitem{kaboli2017pretouch}
Mohsen Kaboli, Di~Feng, Kunpeng Yao, Pablo Lanillos, and Gordon Cheng.
\newblock A tactile-based framework for active object learning and
  discrimination using multimodal robotic skin.
\newblock {\em {IEEE} Robotics Autom. Lett.}, 2(4):2143--2150, 2017.

\bibitem{airobot2019}
Tao Chen, Anthony Simeonov, and Pulkit Agrawal.
\newblock {AIRobot}.
\newblock \url{https://github.com/Improbable-AI/airobot}, 2019.

\bibitem{tippur2022design}
Megha~H. Tippur.
\newblock Design and manufacturing methods for a curved all-around camera-based
  tactile sensor.
\newblock page~69, 2020.

\bibitem{ahn2020robel}
Michael Ahn, Henry Zhu, Kristian Hartikainen, Hugo Ponte, Abhishek Gupta,
  Sergey Levine, and Vikash Kumar.
\newblock Robel: Robotics benchmarks for learning with low-cost robots.
\newblock In {\em Conference on robot learning}, pages 1300--1313. PMLR, 2020.

\bibitem{sun2022soft}
Huanbo Sun, Katherine~J Kuchenbecker, and Georg Martius.
\newblock A soft thumb-sized vision-based sensor with accurate all-round force
  perception.
\newblock {\em Nature Machine Intelligence}, 4(2):135--145, 2022.

\bibitem{lloyd82kmeans}
S.~Lloyd.
\newblock Least squares quantization in pcm.
\newblock {\em IEEE Transactions on Information Theory}, 28(2):129--137, 1982.

\bibitem{Hadsell2006siamese}
R.~Hadsell, S.~Chopra, and Y.~LeCun.
\newblock Dimensionality reduction by learning an invariant mapping.
\newblock In {\em 2006 IEEE Computer Society Conference on Computer Vision and
  Pattern Recognition (CVPR'06)}, volume~2, pages 1735--1742, 2006.

\bibitem{zhang21contrastivetactile}
Qiang Zhang, Yunzhu Li, Yiyue Luo, Wan Shou, Michael Foshey, Junchi Yan,
  Joshua~B. Tenenbaum, Wojciech Matusik, and Antonio Torralba.
\newblock Dynamic modeling of hand-object interactions via tactile sensing.
\newblock In {\em {IEEE/RSJ} International Conference on Intelligent Robots and
  Systems, {IROS} 2021, Prague, Czech Republic, September 27 - Oct. 1, 2021},
  pages 2874--2881. {IEEE}, 2021.

\bibitem{chen20simclr}
Ting Chen, Simon Kornblith, Mohammad Norouzi, and Geoffrey~E. Hinton.
\newblock A simple framework for contrastive learning of visual
  representations.
\newblock {\em CoRR}, abs/2002.05709, 2020.

\bibitem{chen2022system}
Tao Chen, Jie Xu, and Pulkit Agrawal.
\newblock A system for general in-hand object re-orientation.
\newblock In {\em Conference on Robot Learning}, pages 297--307. PMLR, 2022.

\bibitem{duan21dexterous}
Haonan Duan, Peng Wang, Yayu Huang, Guangyun Xu, Wei Wei, and Xiaofei Shen.
\newblock Robotics dexterous grasping: The methods based on point cloud and
  deep learning.
\newblock {\em Frontiers Neurorobotics}, 15:658280, 2021.

\bibitem{chen2022visual}
Tao Chen, Megha Tippur, Siyang Wu, Vikash Kumar, Edward Adelson, and Pulkit
  Agrawal.
\newblock Visual dexterity: In-hand dexterous manipulation from depth.
\newblock {\em arXiv preprint arXiv:2211.11744}, 2022.

\bibitem{downs2022google}
Laura Downs, Anthony Francis, Nate Koenig, Brandon Kinman, Ryan Hickman, Krista
  Reymann, Thomas~B McHugh, and Vincent Vanhoucke.
\newblock Google scanned objects: A high-quality dataset of 3d scanned
  household items.
\newblock {\em arXiv preprint arXiv:2204.11918}, 2022.

\bibitem{shapenet2015}
Angel~X. Chang, Thomas Funkhouser, Leonidas Guibas, Pat Hanrahan, Qixing Huang,
  Zimo Li, Silvio Savarese, Manolis Savva, Shuran Song, Hao Su, Jianxiong Xiao,
  Li~Yi, and Fisher Yu.
\newblock {ShapeNet: An Information-Rich 3D Model Repository}.
\newblock Technical Report arXiv:1512.03012 [cs.GR], Stanford University ---
  Princeton University --- Toyota Technological Institute at Chicago, 2015.

\bibitem{schroff2015facenet}
Florian Schroff, Dmitry Kalenichenko, and James Philbin.
\newblock Facenet: A unified embedding for face recognition and clustering.
\newblock In {\em Proceedings of the IEEE conference on computer vision and
  pattern recognition}, pages 815--823, 2015.

\bibitem{chopra2005learning}
Sumit Chopra, Raia Hadsell, and Yann LeCun.
\newblock Learning a similarity metric discriminatively, with application to
  face verification.
\newblock In {\em 2005 IEEE Computer Society Conference on Computer Vision and
  Pattern Recognition (CVPR'05)}, volume~1, pages 539--546. IEEE, 2005.

\end{thebibliography}
\end{document}